\documentclass[runningheads]{llncs}
\usepackage[T1]{fontenc}
\usepackage{graphicx}

\usepackage{hyperref}
\usepackage{amsmath}
\usepackage{color}

\newcommand{\rec}{\mathit{rec}}
\newcommand{\precision}{\mathit{prec}}
\newcommand{\recg}{\mathit{recG}}
\newcommand{\precg}{\mathit{precG}}
\newcommand{\br}{\mathit{br}}
\newcommand{\tp}{\mathit{TP}}
\newcommand{\fn}{\mathit{FN}}

\usepackage{cleveref}

\usepackage{wrapfig}

\begin{document}
\title{meval: A Statistical Toolbox for Fine-Grained Model Performance Analysis}
\author{Dishantkumar Sutariya\inst{1} \and
Eike Petersen\inst{1}\orcidID{0000-0003-0097-3868} }
%index{Sutariya, Dishantkumar}
%index{Petersen, Eike}
\authorrunning{D. Sutariya et al.}
\institute{Fraunhofer Institute for Digital Medicine MEVIS, Germany\\
\email{eike.petersen@mevis.fraunhofer.de}}
\maketitle              % typeset the header of the contribution
\begin{abstract}
Analyzing machine learning model performance stratified by patient and recording properties is becoming the accepted norm and often yields crucial insights about important model failure modes.
Performing such analyses in a statistically rigorous manner is non-trivial, however.
Appropriate performance metrics must be selected that allow for valid comparisons between groups of different sample sizes and base rates; 
metric uncertainty must be determined and multiple comparisons be corrected for, in order to assess whether any observed differences may be purely due to chance;
and in the case of intersectional analyses, mechanisms must be implemented to find the most `interesting' subgroups within combinatorially many subgroup combinations.
We here present a statistical toolbox that addresses these challenges and enables practitioners to easily yet rigorously assess their models for potential subgroup performance disparities.
While broadly applicable, the toolbox is specifically designed for medical imaging applications.
The analyses provided by the toolbox are illustrated in two case studies, one in skin lesion malignancy classification on the ISIC2020 dataset and one in chest X-ray-based disease classification on the MIMIC-CXR dataset.
\end{abstract}
\section{Introduction}
It is increasingly well-recognized that developers should assess the performance of their machine learning models not just in the aggregate, but also dis-aggregated -- or \emph{stratified} -- by attributes characterizing model inputs and targets~\cite{OakdenRayner2020a,Collins2024,Petersen2023}.
The motivation for such stratified analyses is twofold.
Firstly, they enable the identification of \emph{quality-of-service (QoS) biases}, i.e., whether models perform differently well in different patient cohorts.
Secondly, and more broadly, such fine-grained analyses can reveal important model failure modes, such as shortcut learning or failures on specific recording devices, field strengths, protocols, or other technical parameters~\cite{Lotter2024,JimenezSanchez2023}.
Such shortcomings are often not readily apparent in aggregate test-set performance analyses, and they can also be a cause of QoS biases~\cite{Olesen2024}.
In essence, finely stratified performance analyses allow answering the crucial question: \emph{Does this model work for every patient?}%\footnote{Notably, the EU's AI Act separately demands both checking for bias with regards to sensitive attributes as well as investigating overall model robustness, recognizing the importance of fine-grained performance evaluation.}

Performing such stratified model performance analyses in a statistically rigorous manner is far from trivial, however.
Subgroups may differ strongly in sample size and base rates (incidences), rendering standard performance assessment tools such as precision-recall (PR) curves (and the area below them) as well as the expected calibration error (ECE) and its variants inapplicable~\cite{Petersen2023}.
To assess the uncertainty of performance evaluations in subgroups, the computation of reliable uncertainty estimates (confidence intervals) is indispensible, and appropriate statistical tests must be designed while accounting for the fact that potentially \emph{many} subgroups are analyzed and there is, thus, a high risk of false positive findings.
Much of the existing literature on performance evaluation methodology focuses on the case of aggregate evaluation~\cite{Collins2024}, providing little guidance on proper methodology for stratified analyses.
While there is a large branch of literature on comparative model evaluation~\cite{Raschka2018,Varoquaux2023,Rainio2024}, these works mostly focus on the case of \emph{model comparison} which differs crucially: in that setting, the samples of interest (the same test set analyzed using different models) are \emph{paired}, with important implications for proper statistical methodology.

Here, we present a statistical toolbox designed to address these challenges. Our aim is to to empower medical imaging practitioners to rigorously assess their models with respect to intersectional subgroup performance disparities.

\section{Related work}
Prior work has investigated the limited applicability of commonly-used metrics.
The precision-recall (PR) curve, often recommended in the case of strong class imbalance and widely used in the medical domain~\cite{Varoquaux2023}, depends on the base rate $p(y=1)$ of the sample under test~\cite{Flach2015,Boyd2012,Varoquaux2023} and is therefore not meaningfully comparable between samples with different base rates~\cite{Petersen2023}.\footnote{This issue does not affect the AUROC metric, which is base rate-independent.}
This also affects derived metrics such as the area under the precision-recall curve (AUPR), sometimes also called Average Precision (AP), and the geometric mean of precision and recall, the $F_1$ score.
Several alternative base rate-independent metrics have been proposed~\cite{Flach2015,Boyd2012}.
In terms of calibration measurement, the expected calibration error (ECE) -- the most commonly used metric -- suffers from a strong sample size bias~\cite{Broecker2011,Roelofs2022,Kumar2019,Gruber2022}.
Its value for an identically calibrated model thus changes as a function of the test sample size, rendering this metric ill-suited for subgroup comparisons~\cite{Petersen2023,RicciLara2023}.
Debiased alternative metrics have been proposed~\cite{Kumar2019,Ferro2012,Roelofs2022}.

Addressing the challenge of identifying potential biases in combinatorially many intersectional subgroups, Kearns et al.~\cite{Kearns2017} and Zhang et al.~\cite{Zhang2017a} provide efficient algorithms for specific performance metrics. 
The extension of these approaches to commonly used metrics such as AUROC is not obvious, however.
Most closely related to our work, Cherian et al.~\cite{Cherian2024} recently note that proper statistical testing in the fairness auditing scenario is under-addressed; they develop a comprehensive and rigorous statistical approach to certifying subpopulation performance disparities. Their approach is highly methodical in nature and deviates from standard model evaluation workflows, limiting accessibility to medical imaging practitioners.
Finally, DiCoccio et al.~\cite{DiCiccio2020} describe a generalized approach to statistical hypothesis testing for arbitrary metrics in the fairness auditing case; we implement this approach in our toolbox.

Toolboxes such as AIF360~\cite{Bellamy2018} and fairlearn~\cite{Weerts2023a} may appear to provide similar functionality, but they do not address the specific needs of comprehensive intersectional model evaluation: many standard model performance metrics are not available, and neither statistical testing methodology nor stratified analyses of common performance curves (ROC, PR, calibration) are provided.

\section{Methodology}
\subsection{Metric choices.}
We implement standard performance metrics including (balanced) accuracy, AUROC, (balanced) Brier score, sensitivity and specificity.
Any metric that is an average over per-recording metric values (such as the average dice score) is implemented via a blanket `AverageMetric', providing full support in terms of confidence intervals and statistical testing.
In addition, we also provide implementations of several non-standard metrics in the toolbox. 
Most notably, we implement the (partial) area under the precision-recall-gain curve (pAUPRG), originally proposed by Flach et al.~\cite{Flach2015} and the debiased root mean squared calibration error (DRMSCE) proposed by Petersen et al.~\cite{Petersen2023}, which represents an improved version of the debiased estimator proposed by Kumar et al.~\cite{Kumar2019}.

The AUPRG metric was originally proposed by Flach et al.~\cite{Flach2015} to address several noted deficiencies of the AUPR (or AP) metric, including its base rate dependence. 
Flach et al. define the precision and recall \emph{gains} as %harmonically base rate-rescaled versions of precision and recall, namely,
\begin{equation}
    \precg = \frac{\precision - \br}{(1-\br)\precision} \qquad \text{and} \qquad \recg = \frac{\rec - \br}{(1-\br)\rec}
\end{equation}
where $\br=P(y=1)$ denotes the base rate of the test sample.
Flach et al. also provided an implementation of their proposed metrics and the area below PRG curves.\footnote{\url{https://github.com/meeliskull/prg}}
This implementation has been unmaintained for many years, however, and suffers from several long-known issues, warranting an up-to-date reimplementation.
In addition, we discovered a previously undescribed problem with the original method of calculating AUPRG, which we will describe in the following.

The AUPRG is obtained by integrating over the PRG curve from $\recg = 0$ to $\recg=1$.
$\recg=0$ corresponds to $\rec = \br$, so this integration requires there to be a well-defined $(\recg, \precg)$ point at $\rec=\br$.
If there is no decision threshold that happens to yield exactly $\rec=\br$, this point can be obtained by linear interpolation (which is meaningful in PRG space, unlike in PR space~\cite{Flach2015}) if and only if there are well-defined points on either side of $\rec=\br$.
`Well-defined' here refers, in particular, to the value of $\precision$, which is only defined if there is at least one positive prediction.
The smallest $\rec$ value for which $\precision$ is well-defined is thus given by the decision threshold corresponding to the highest score value predicted by the model.
If the highest predicted score is for a \emph{negative} example, this yields a valid point at $\rec=0$ and $\precision=0$, and we can thus obtain the PRG point at $\recg=0$ by linear interpolation.
%a decision threshold which yields well-defined values for $\rec$ and $\precision$ with $\rec \leq \br$.\footnote{Unlike PR space, linear interpolation in PRG space is valid~\cite{Flach2015}: a point at $\recg=0$ can be obtained by linear interpolation if the neighboring points for both $\recg < 0$ and $\recg > 0$ are associated with well-defined precision values. The latter condition (valid point for $\recg > 0$) is usually satisfied, the first condition is the critical one.} 
%This is satisfied if the highest predicted model score is for a negative example, yielding a valid point at $\rec=0$ and $\precision=0$.
Alternatively (and more likely), if there is at least one \emph{positive} example in the set of samples obtaining the highest score, we obtain the first well-defined point at
\begin{equation}
    \rec = \frac{\tp}{\tp+\fn} = \frac{\text{num positives at highest score}}{\text{all positives}}.
\end{equation}
Especially for small samples sizes and strong class imbalance ($\br \ll 0.5$), this point will often be at $\rec > \br$, rendering AUPRG ill-defined.
For this reason, we also provide an implementation of a \emph{partial} AUPRG that is obtained by integrating $\precg$ over $[\recg_{\text{min}}, 1]$ with $0 \leq \recg_{\text{min}} < 1$.

For all metrics derived from an underlying \emph{curve}, such as AUROC, AUPR(G), and DRMSCE (derived from the calibration curve), we also present the corresponding curves split by groups, and confidence intervals obtained by bootstrapping~\cite{Austin2014}.
The operating points selected by a given decision threshold for the different groups are highlighted.

\subsection{Intersectional analyses}
Performance disparities should not only be evaluated between groups defined by a single attribute (gender) but also between groups defined by the intersection of multiple attributes (gender $\times$ age $\times$ technical parameters $\times$ ...)~\cite{Kearns2017}.
This presents its own challenges: there are combinatorially many subgroups to consider, some of which will be very small (further increasing both the importance and the difficulty of valid metric uncertainty quantification).
Taking a pragmatic approach, we simply allow the user to set a minimum group membership threshold for a subgroup to be considered, as well as a maximum interaction level of attributes.
For visualization purposes, we select the most `interesting' subgroups to display based on the sum of a group's ranks in terms of the p-value attached to its performance disparity and the magnitude of that performance disparity, inspired by the `volcano plots' often used for similar purposes~\cite{Li2014}.

\subsection{Confidence intervals}
All metric results are accompanied by associated confidence intervals (CIs) quantifying the uncertainty about the population-level metric value caused by the fact that the model is evaluated on a small sample drawn from the overall population.
For any metric, CIs can be obtained in one of two ways.
Firstly, analytical CI approximations can be implemented if they are available for a given metric. 
In the current version, we provide analytical CIs for AUROC\footnote{Fast DeLong's method~\cite{Sun2014} as implemented in the `confidenceintervals' package~\cite{Gildenblat2023} and a custom implementation of Newcombe's method~\cite{Newcombe2005} for small groups ($\leq 50$ samples) and groups with perfect separation (AUROC = 1.0) since DeLong is known to provide very poor coverage for these cases~\cite{Feng2015}.} and ratio-based metrics (Accuracy, Sensitivity, etc.; we use the Wilson score interval implementation of statsmodels~\cite{Seabold2010}).
In addition, CIs for any metric can be obtained using a standard percentile bootstrap.
In the case of metrics requiring both positive and negative samples (e.g., AUROC) and few samples, we stratify the bootstrap to prevent the excessive occurrence of undefined metric values.

\subsection{Statistical hypothesis testing}
In the case of a single binary attribute of interest (say, gender), it is clear which statistical hypotheses to test for: does model performance differ significantly between these two groups?
It is less clear which question to ask (and which hypotheses to test) in the case of subgroups defined by the combination of multiple categorical attributes.
Simply testing for pairwise differences between all subgroups results in combinatorially many tests, implying a need to correct for an equally large number of multiple tests and, thus, a high chance of null results even in the presence of non-negligible performance differences.
In addition, we are also interested in subgroups of different cardinalities, i.e., groups defined by a different number of attributes. Is it meaningful to compare model performance between, say, `women' and `young women'?
Finally, choosing specific subgroups to test for significant differences \emph{after} inspecting the results of the model performance assessment would constitute HARKing: `hypothesizing after the results are known,' a common malpractice closely related to p-hacking~\cite{Kerr1998,Stefan2023}.

In order to circumvent all of these issues, we propose to test for differences with respect to each subgroup's \emph{complementary} group, defined as assuming different values for each group-defining attribute. 
For instance, if the group under test were defined as `gender $=$ female and age $<$ 25', we would test for differences with respect to the group `gender not female and age $\geq$ 25'.
This approach significantly reduces the number of tests to perform compared to the pairwise approach, while still allowing for an exploratory analysis.

To provide a general method for statistical significance testing that is valid for any metric, we implement the approach proposed by DiCiccio et al.~\cite{DiCiccio2020}, in essence a permutation-based test using a studentization of the metric of interest.
The studentization requires an estimate of the variance of a given metric's value on a given (permuted) dataset.
This variance can be obtained via bootstrapping, but this approach is computationally expensive as it must be repeated for every permutation.
We therefore (as also suggested by DiCiccio et al.) use analytical expressions for the variance of a metric wherever they are available.
We correct for multiple hypothesis testing using the Holm--Bonferroni correction.

\subsection{Implementation}
The toolbox is implemented in python and designed to be modular, easily extensible, and easy to use.
It is publicly available, including all code required to reproduce the two case studies presented below.\footnote{\url{https://github.com/FraunhoferMEVIS/meval}}
Visualizations are created using the plotly library, which enables both interactive visualizations and static figure exports.
As inputs, the library requires a pandas dataframe with model predictions, ground truth information, and any available metadata.
In addition, the user must specify the metrics to analyze.
No access to the model is required.
Using a single function call, the library creates an interactive HTML report that summarizes model performance across intersectional subgroups.
The results are also returned in raw form to enable further custom analyses.

\section{Case studies}
We present two case studies to illustrate the kinds of analyses enabled by our toolbox.
All plots presented in the following represent direct outputs of simple toolbox function calls, with no further customization applied.

\subsection{ISIC skin lesion malignancy classification}
We use the training split of the ISIC2020 dataset~\cite{Rotemberg2021}. We remove duplicate images based on the list provided on the dataset website and split the data into an 80\% training set and a 20\% evaluation split, ensuring no lesion leakage based on the 'lesion\_id' metadata field.\footnote{Patient leakage can still occur~\cite{Cassidy2022}; we deemed this to be non-critical for our study.}
The images are resized to $256 \times 256$ pixels, center-cropped to $224 \times 224$ pixels, and normalized using the parameters provided in the torchvision documentation for the ImageNet-pretrained ResNet50-V2.
We finetune the model for 25 epochs using stochastic gradient descent (binary cross-entropy, learning rate $5\times 10^{-4}$, momentum $0.9$, batch size 64), random flips and random color jitter (torchvision 0.22.1, all parameters set to 0.25).
%The model obtains AUROC 0.86 on the evaluation split.
%All further analyses presented are obtained on the test set, with subgroups defined by sex (binary), anatomical site, and age group.

\begin{figure}[p]
    \includegraphics[width=\textwidth]{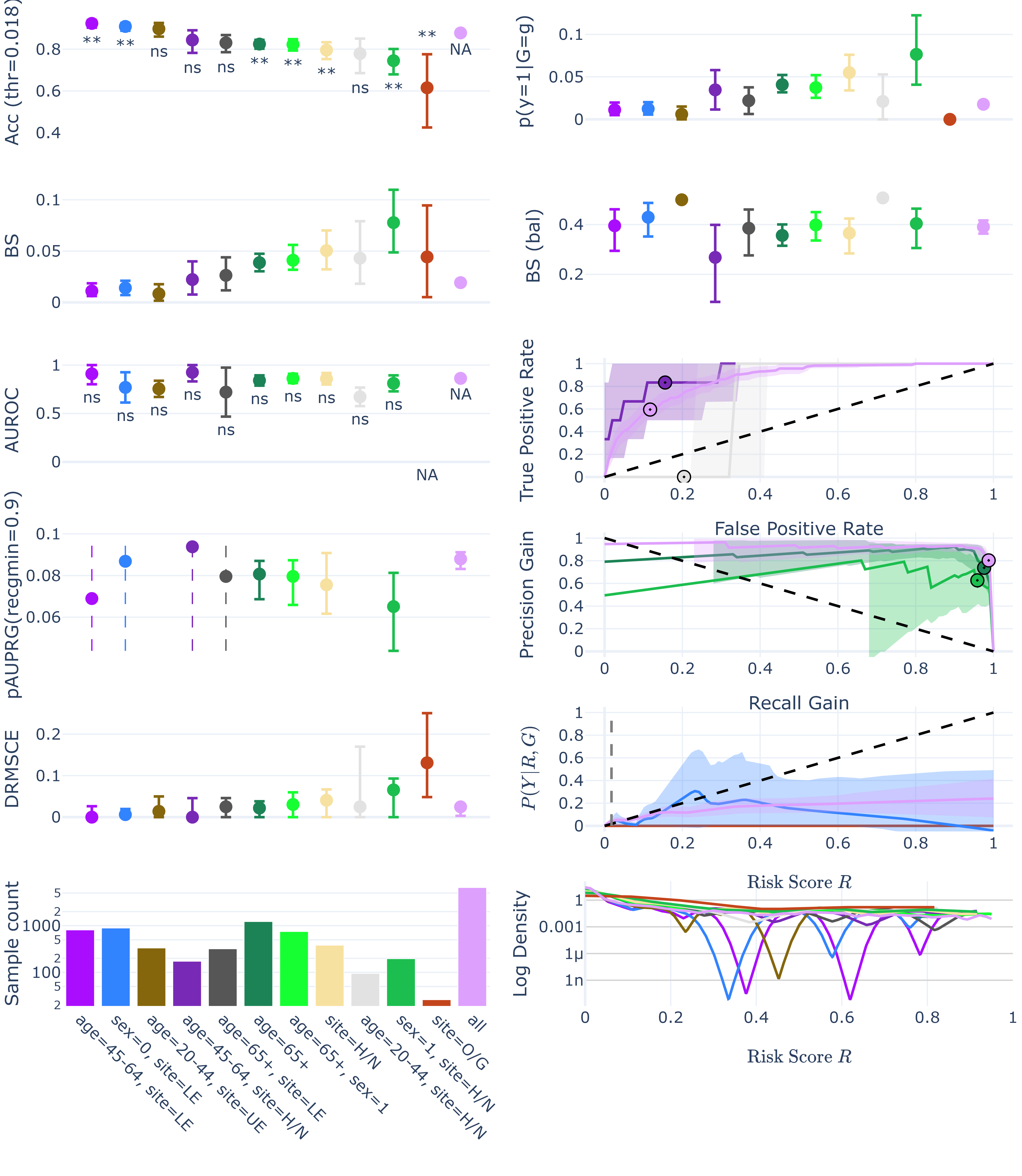}
    \caption{Exemplary default output of the toolbox for the ISIC2020 case study. Statistical tests performed for accuracy and AUROC. For thresholded metrics, the base rate was chosen as the decision threshold. H/N: head/neck, LE: lower extremity, UE: upper extremity, O/G: oral/genital, TO: torso. ns: not significant ($p > 0.01$), *: $p \leq 0.01$, **: $p \leq 0.001$. Dashed vertical lines indicate that CIs could not be obtained.}
    \label{fig:isic}
\end{figure}

\Cref{fig:isic} shows a (static version of the otherwise dynamic) metric overview generated using our toolbox.
Classification accuracy differs significantly between subgroups, but this is apparently primarily a function of the respective subgroups' base rates: no statistically significant AUROC differences are found. We also observe that the model is very poorly calibrated overall.

\subsection{MIMIC-CXR lung disease diagnosis}
We use the MIMIC-CXR-JPG database~\cite{Johnson2024,Johnson2019,Goldberger2000}, discarding lateral recordings and keeping only frontal (AP/PA) recordings.
We discard the `support device', `fracture' and `pleural other' labels, focusing our analyses on the remaining 10 disease labels and the `No finding' label.
Following the approach of Weng et al.~\cite{Weng2023}, we discard multiple recordings for the same patient and keep just one out of the set with the most disease labels, in order to minimize the risk of label errors~\cite{Zhang2022}.
From the resulting dataset of 41,168 recordings, a test set is constructed by randomly sampling 35 positive instances for each of the 11 labels for each of the top-5 race groups, resulting in a total test-set size of 1,757 samples.\footnote{For some race--label combinations, less than 35 samples were available. In those instances, we used all available samples in the test set. Notice also that due to the multi-label nature of the data, there may be more than 35 samples from a given race-label combination in the resulting test set.}
The remaining data are randomly split into a training and validation set of 37,439 (95\%) and 1,972 (5\%) samples, respectively. We ensure that there is no patient overlap between any of the three sets.
We fine-tune a DenseNet121 for multilabel classification, similar to prior work~\cite{SeyyedKalantari2020,Yang2024a}.

\begin{figure}[t]
    \includegraphics[width=0.47\textwidth]{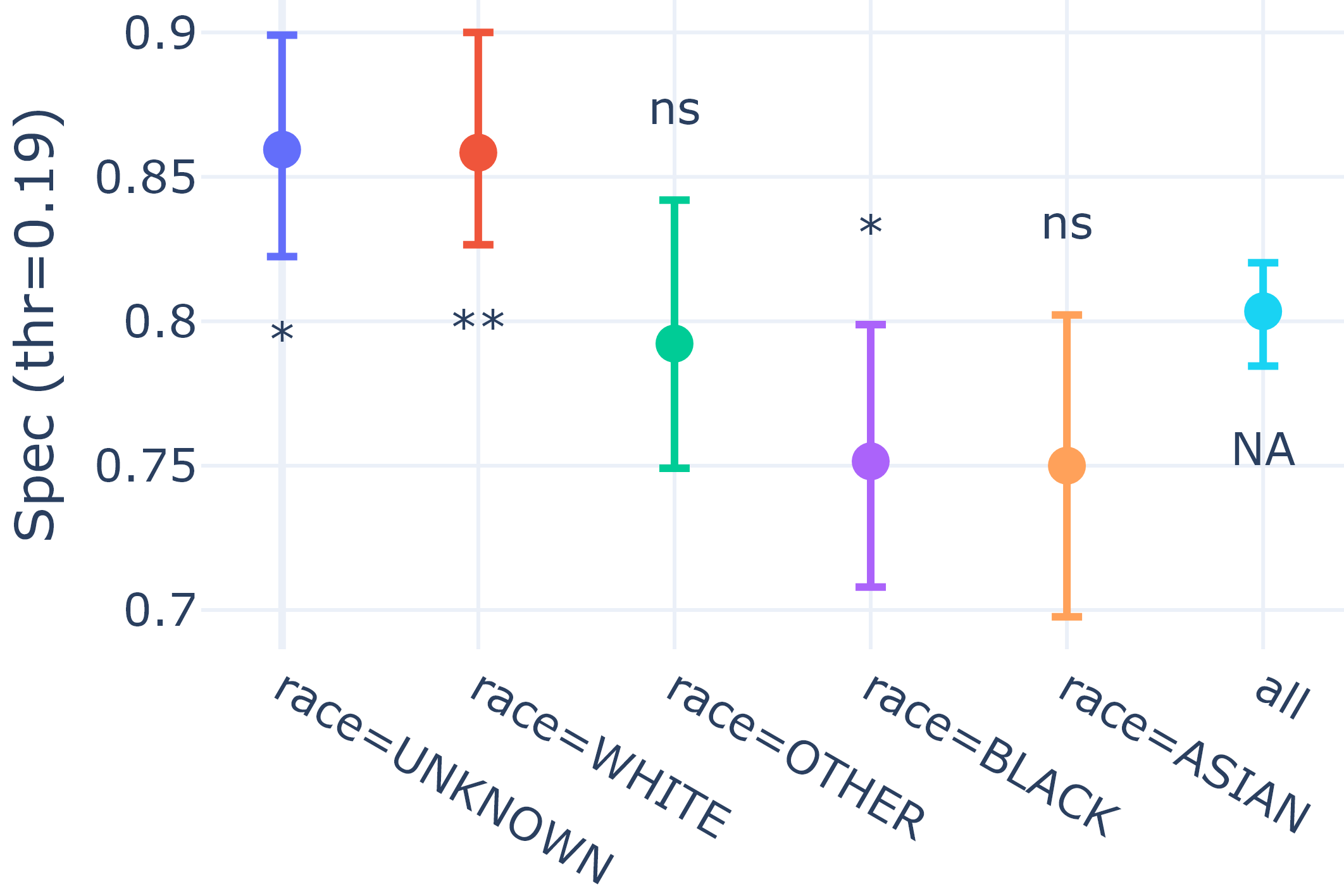}%
    \hfill
    \includegraphics[width=0.47\textwidth]{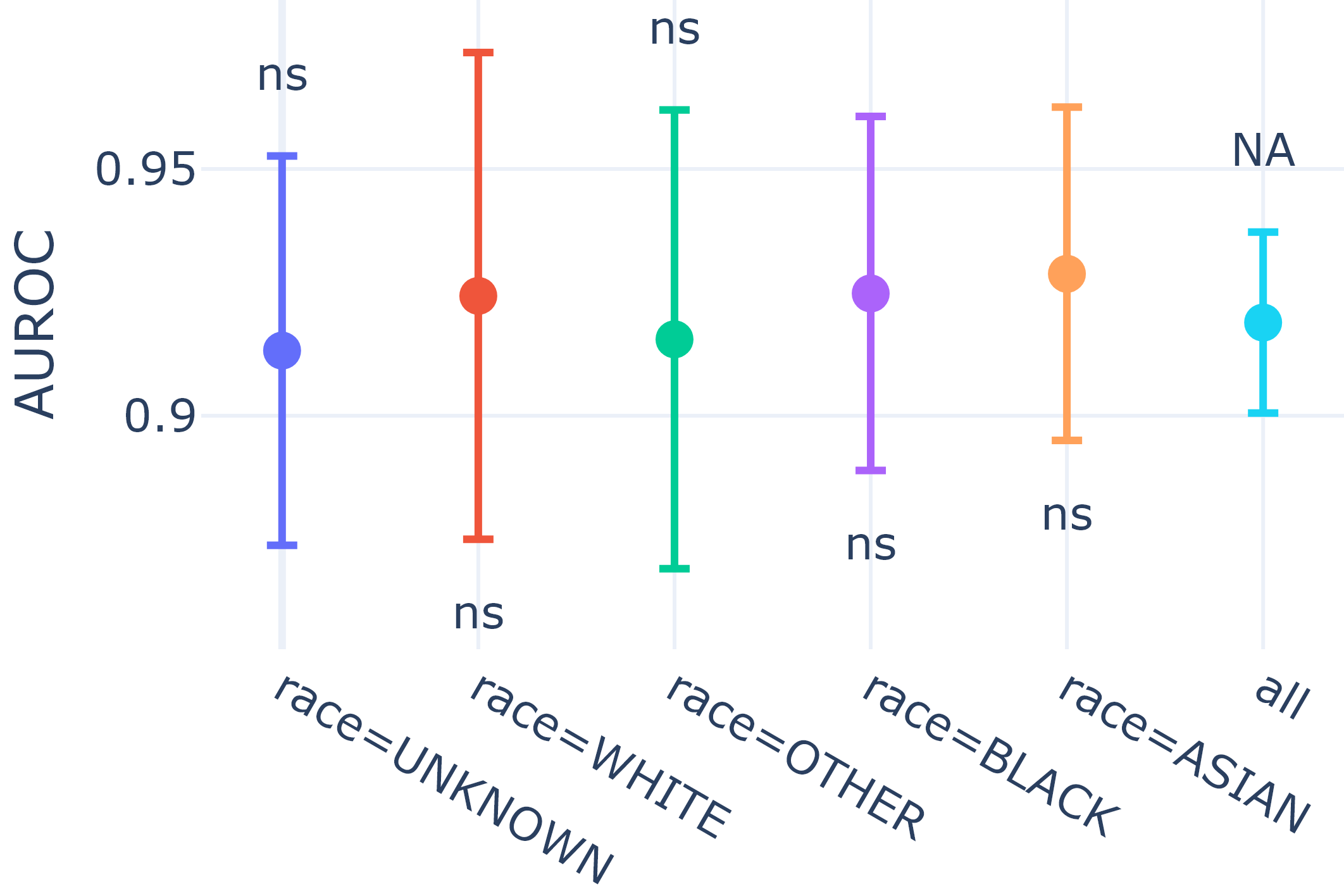}%
    \caption{No-Finding Specificity (left) and AUROC (right), stratified by racial groups. For specificity, the threshold was chosen as in \cite{SeyyedKalantari2021} to maximize the geometric mean of sensitivity and specificity. ns: not significant ($p > 0.01$), *: $p \leq 0.01$, **: $p \leq 0.001$.}
    \label{fig:mimic}
\end{figure}

\begin{wrapfigure}{r}{0.35\textwidth}
    \includegraphics[width=0.32\textwidth]{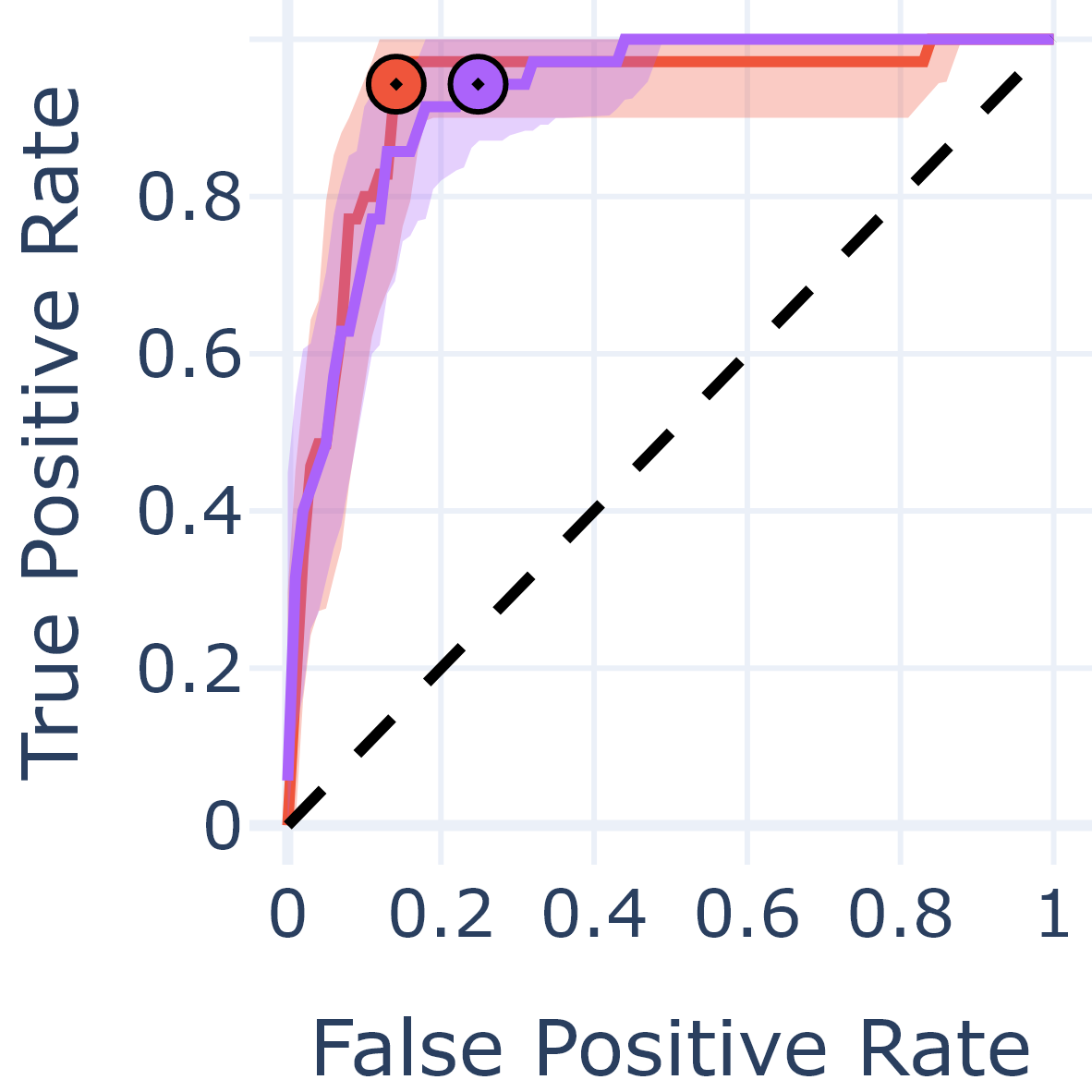}
  \caption{ROC curves and operating points for white (red) and black (purple) patients.}
  \label{fig:mimic-roc}
  %\vspace{-0cm}
\end{wrapfigure}

As preprocessing steps, the data are downscaled to $224 \times 224$ pixels, randomly rotated (max. $\pm 10^\circ$), randomly cropped and resized to $200 \times 200$ pixels and randomly flipped (horizontally and vertically).
We optimize binary cross-entropy using AdamW with early stopping (patience 5) based on the validation AUROC. 
The best (in terms of validation AUROC) intermediate checkpoint is retained, which achieves a macro-averaged validation AUROC of 0.84, and macro-averaged test AUROC of 0.79.
Notice that our test set is specifically constructed to be highly diverse and challenging, and a drop in AUROC between the training/validation and test sets is thus to be expected.
For our subgroup analyses, following Seyyed-Kalantari et al.~\cite{SeyyedKalantari2021}, we focus on the 'No finding' label and racial groups. \Cref{fig:mimic} shows the per-subgroup `No-finding' FPR and AUROC.

We reproduce the finding of Seyyed-Kalantari et al. that there is a significant gap in the 'No Finding' specificity between racial groups.
Interestingly, however, we observe that there is no significant difference between these groups in terms of AUROC. A comparison of the ROC curves (\cref{fig:mimic-roc}) indicates that while the overall ROC curves (and the areas under them) are similar, the racial groups are in different ROC operating points (TPR/FPR) for the same decision threshold. %, suggesting an issue related to model miscalibration as the root cause of the apparent 'No Finding' disparity.

\section{Conclusion and outlook}
Our aim with this work is to provide a statistical toolbox that enables practitioners to conduct rigorous intersectional performance disparity analyses.
To this end, we implement several non-standard performance metrics, best-practice statistical methodology, and interactive visualizations for exploring potential disparities.
Future work may include the development of a deconfounding approach for disentangling the effects of different causal factors on model performance~\cite{Pfohl2025,Mukherjee2022,Petersen2023a}.
We hope that the publication of our toolbox may inspire many researchers to perform case studies in different fields of application, aiding the identification of model blind spots and unfair biases.

\begin{credits}
\subsubsection{\ackname} 
The authors would like to thank Dr. Max Westphal and M.Sc. David Pfrang for helpful discussions on the statistical methodology.
Part of the work that led to this publication has received funding from the European Union’s Horizon Europe research and innovation programme under grant agreement No 101057091.

\subsubsection{\discintname}
The authors declare no competing interests.

\end{credits}

\bibliographystyle{splncs04}
\bibliography{references}

\end{document}